\documentclass[10pt,twocolumn,letterpaper]{article}

\usepackage[pagenumbers]{iccv} 

\usepackage{booktabs}
\usepackage{caption}

\definecolor{iccvblue}{rgb}{0.21,0.49,0.74}
\usepackage[pagebackref,breaklinks,colorlinks,allcolors=iccvblue]{hyperref}
\usepackage{tipa}

\def\paperID{19} 
\def\confName{ICCV}
\def\confYear{2025}

\title{PIA: Deepfake Detection Using Phoneme-Temporal and Identity-Dynamic Analysis}

\author{
Soumyya Kanti Datta\thanks{Equal contribution.}, Tanvi Ranga\footnotemark[1], Chengzhe Sun, Siwei Lyu\\
University at Buffalo, SUNY\\
Buffalo, NY, USA\\
{\tt\small \{soumyyak, tanviran, csun22, siweilyu\}@buffalo.edu}
}

\begin{document}
\maketitle
\begin{abstract}
The rise of manipulated media has made deepfakes a particularly insidious threat, involving various generative manipulations such as lip-sync modifications, face-swaps, and avatar-driven facial synthesis. Conventional detection methods, which predominantly depend on manually designed phoneme–viseme alignment thresholds, fundamental frame-level consistency checks, or a unimodal detection strategy, inadequately identify modern-day deepfakes generated by advanced generative models such as GANs, diffusion models, and neural rendering techniques.
These advanced techniques generate nearly perfect individual frames yet inadvertently create minor temporal discrepancies frequently overlooked by traditional detectors. We present a novel multimodal audio-visual framework, Phoneme-Temporal and Identity-Dynamic Analysis(PIA), incorporating language, dynamic face motion, and facial identification cues to address these limitations. We utilize phoneme sequences, lip geometry data, and advanced facial identity embeddings. This integrated method significantly improves the detection of subtle deepfake alterations by identifying inconsistencies across multiple complementary modalities. Code is
available at \url{https://github.com/skrantidatta/PIA}
\end{abstract}
\section{Introduction}
\label{sec:intro}

The rapid advancement of generative AI technologies has resulted in an increase in tools capable of creating synthetic media. These tools, in turn, have led to a proliferation of deepfakes, which are AI-created or manipulated videos. The distinctions between authentic and deepfake media become increasingly challenging for human viewers. Although deepfakes involving humans have gained significance in the entertainment sector, they also pose a significant risk to identity, trust, and societal integrity. Recent events have highlighted the potential identity and security vulnerabilities posed by deepfakes. In February 2024, a multinational corporation incurred a loss of \$25 million due to an employee being deceived by a deepfake impersonation of their chief financial officer and other senior officials. The employee, perceiving it as a legitimate request, transferred funds to a fake account \cite{ChenMagramo2024}. In another incident, a deepfake impersonator from North Korea deceived KnowBe4, a cybersecurity firm, into employing them in the latter half of 2024 \cite{Bracken2024}. The ability to deceive a cybersecurity firm demonstrates the remarkable efficacy of these forgeries.

\begin{figure*}[t]
    \centering
    \includegraphics[width=1.0\textwidth]{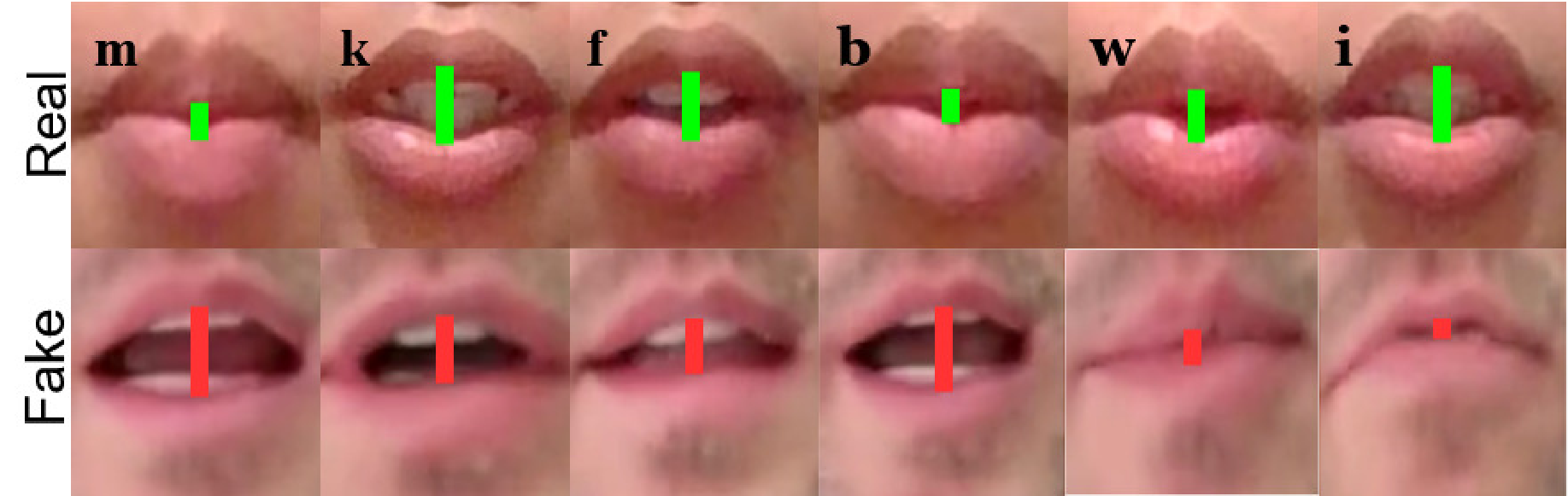} 
    \caption{Comparison of lip shapes corresponding to different phonemes. Note that in lip-sync deepfake videos, the degree of lip closure often does not correspond to the sound being pronounced.}
    \label{fig:main_fig} 
    \vspace{-2mm}
\end{figure*}


Most existing deepfake detection methods use only one modality, predominantly relying on analysis focused solely on audio or visual signals. There are several works, such as   \cite{Agarwal_2020_CVPR_Workshops}, 
and  \cite{krause2023language}, in which they use multimodal cues but rely on rule-based alignment for audio-visual cues. However, such methods are insufficient in identifying complex manipulations generated by recent developments in generative adversarial networks (GANs) \cite{goodfellow2014generative} and diffusion-based \cite{ho2020denoising,dhariwal2021diffusion} models, as these models generate high-fidelity facial dynamics and speech-driven articulations that reduce conventional audio-visual alignment discrepancies.

In this work, we develop a new multimodal deepfake detection method, Phoneme-Temporal and Identity-Dynamic
Analysis (PIA), to identify audio-visual inconsistency patterns using temporal inconsistencies cues between audio and visual signals. Prior work by Agarwal et al.~\cite{Agarwal_2020_CVPR_Workshops} has demonstrated that such phoneme-viseme mismatches can serve as reliable indicators of manipulation, particularly in the context of deepfakes generated by automated lip-syncing.

We hypothesize that deepfakes insufficiently replicate the sophisticated visemic articulation corresponding to particular phonemes, especially bilabial and rounded vowels such as /m/,/b/,/p/, and /o/. In actual human speech, these phonemes correspond to distinct and reproducible articulatory motions, including complete lip closures for /m/,/b/, and /p/ and a characteristic lip rounding for /o/. Real video sequences exhibit a regular pattern of lip geometry at frames temporally synchronized with these phonemes, illustrating the physiological limitations of human articulation.
This phenomenon is closely related to the McGurk effect \cite{mcgurk1976hearing}, where conflicting auditory and visual speech cues result in a perceptual illusion. It highlights how strongly the human brain relies on audio-visual coherence, especially in speech perception.
Additionally, we hypothesize that ArcFace \cite{deng2019arcface} embeddings in real videos exhibit a progressive and consistent temporal progression, marked by smooth transitions between successive frames. In contrast, fake videos, especially those produced by face-swapping methods, frequently exhibit sudden and inconsistent shifts in the embedding space. The $\ell_2$ distance, or Euclidean distance, is calculated as the square root of the sum of the squared differences between comparable elements of two embedding vectors. This metric measures the frame-to-frame variation in facial identity representation as conveyed by ArcFace. These data indicate that significant abrupt changes in embedding distance can act as a reliable indicator for potential manipulation in an operational deepfake detection system.

Our approach introduces a multimodal architecture directed by a distinct temporal loss function that correlates 14 diverse spoken phoneme letters with lip shape, mouth closure score, and facial identity variation to identify temporal inconsistencies between audio and various facial landmarks. By combining multimodal signals, the model easily discerns the three primary manipulation media: lip-sync, face-swap, and avatar-based deepfakes. We performed comprehensive experiments and ablation studies on two benchmark datasets to assess the effectiveness of our method. The results demonstrate the robustness of our model in identifying in-domain deepfakes.

In summary, our work presents the following primary contributions:
\begin{itemize}
    \item We highlight the temporal discrepancies between audio-visual cues across 14 distinct phoneme letters, including bilabials and vowels.
    \item We present an innovative deepfake detection pipeline that integrates three complementary streams (viseme images,  identity embeddings, and lip geometry) through a shared multi-headed attention mechanism and controlled fusion.
    \item We systematically measure the effect of each modality in  controlled ablation studies.
    \item Our methodology attains an AUC of 98\% on the DeepSpeak v2.0 dataset.
    \item We propose an auxiliary loss that penalises temporal inconsistencies in identity embeddings across successive frames. This loss promotes a more consistent identity representation over time that helps to detect abrupt identity shifts, often observed in face-swap deepfakes.
\end{itemize}



\section{Related Work}

\noindent{\bf Deepfake generation}. As AI technologies advance, more sophisticated and easily accessible deepfake generation tools have become widespread. Deepfake videos fall into two main categories: entire-face synthesis (e.g., face-swaps or talking‐head generation, such as avatar deepfakes) and partial manipulation, such as lip-syncing deepfakes, which alter only lip movements to match audio. Early video deepfake generation focused on face-swap techniques such as \cite{face-swap, fsgan, facefusion, inswapper,simswap} where the entire face of the original identity is replaced with the target identity. 
Lip syncing deepfakes overwrite only the mouth region to match new audio. \cite{wav2lip} trains a GAN to produce lip movements conditioned on speech features, while \cite{videoretalking} uses a three-stage process: expression stabilizer, a lip-sync model, and a face enhancement model to create talking heads.  More recent models \cite{diff2lip,latentsync} use diffusion‐based models to yield sharper and more temporally coherent lip-syncing deepfakes.
 Avatar deepfakes animate a single image to produce a full talking‐head video to match a target audio or video. \cite{liveportrait, hellomeme,zheng2024memo} use multi‐stage control and diffusion modules to extract identity, motion, voice, and emotion embeddings to produce highly realistic speaking avatars. Such methods have consequently made deepfakes increasingly difficult to detect.

\noindent{\bf Deepfake detection}. With deepfake generation methods becoming more sophisticated, detection techniques have advanced in parallel as well. Existing methods often leverage either audio, video, or multimodal signals, each with distinct approaches to feature extraction, mismatch detection, and classification techniques.

Visual‐only detectors analyze spatial artifacts and temporal inconsistencies in frame sequences \cite{Xception, li2018exposing, nirkin2021deepfake, huang2023implicit, datta2024exposing, haliassos2021lips, zheng2021exploringftcn}. These models aim to detect inconsistencies in the pixel domain,
introduced during the manipulation process. The audio–visual detectors exploit the lip–speech mismatches to detect deepfakes \cite{feng2023self, yu2024explicit, yang2023avoid}. For example, \cite{Agarwal_2020_CVPR_Workshops}  proposed a multimodal method utilizing phoneme–viseme alignment mismatch to detect deepfake videos. \cite{chen2023npvforensics} integrates audio, video, and physiological signals, capturing multimodal discrepancies including lip-sync and physiological anomalies like temperature variations and pulse irregularities to detect deepfakes. \cite{oorloff2024avff} proposed a two-stage video detector that first self-supervises on real videos to learn intrinsic audio–visual correspondences via contrastive and autoencoding objectives, then fine-tunes on real vs. fake data. Further improving on the previous models, \cite{liu2024lipfd} focuses on temporal audio-visual inconsistencies, employing global and local encoders built on Vision Transformer \cite{dosovitskiy2020image} pre-trained on CLIP \cite{radford2021learning} to identify mismatches. Its classifier utilizes transformer architectures equipped with a dynamic attention module to detect deepfakes.




Our proposed pipeline advances beyond these existing methods by integrating a richer set of multimodal inputs, including phonemes from WhisperX \cite{bain2022whisperx} combined with phonemizer, visemes and lip landmarks from MediaPipe \cite{lugaresi2019mediapipe}, and facial identity embeddings from ArcFace \cite{deng2019arcface}. We introduce nuanced mismatch signals such as soft lip-closure scores, viseme-phoneme mismatches, and ArcFace drift to capture subtle articulatory and identity-based discrepancies. This comprehensive approach demonstrates improved performance in intra-dataset performance and cross-manipulation generalization.


\begin{figure}[t]
    \centering
    \includegraphics[width=\columnwidth]{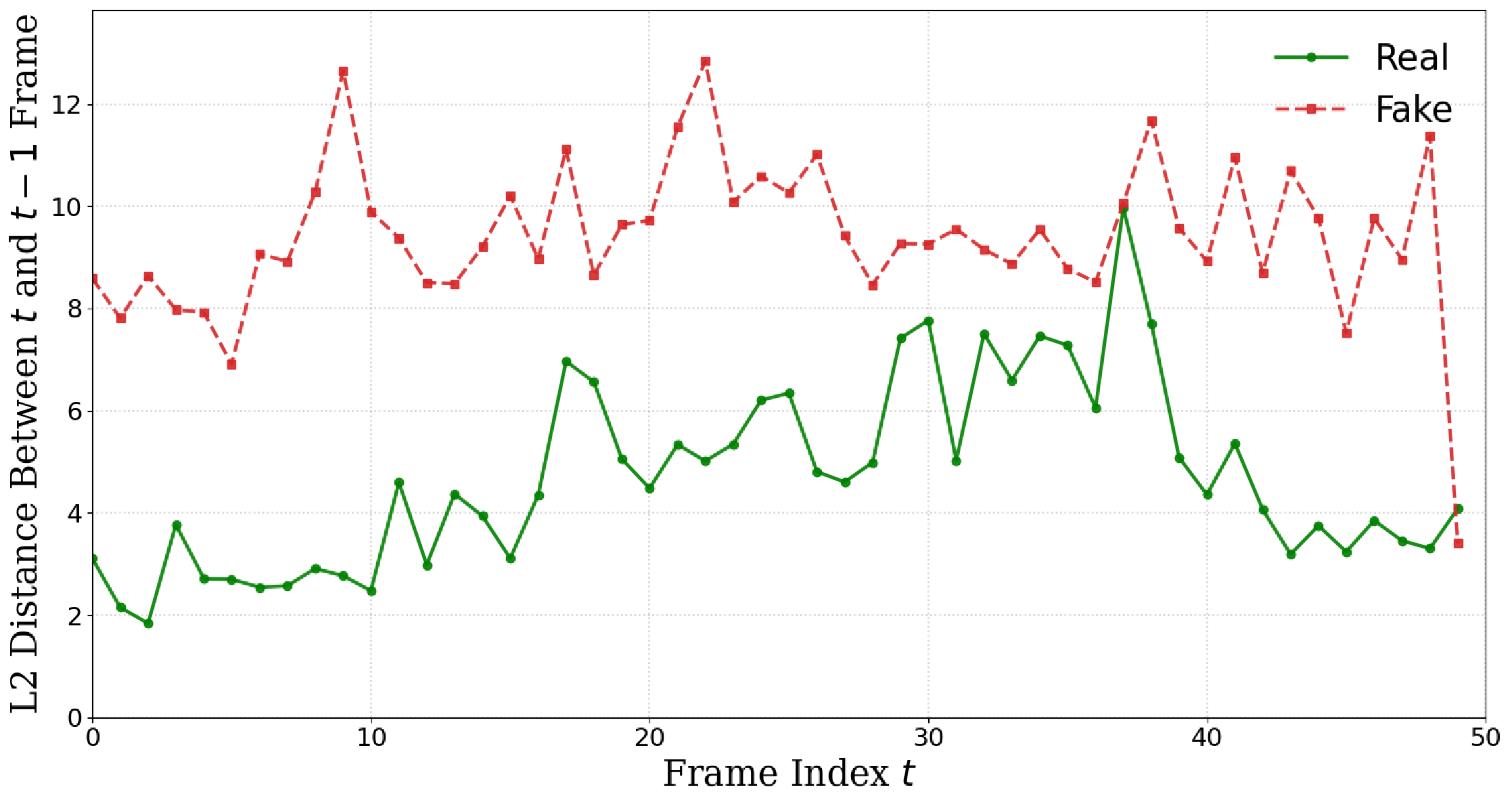}
    \caption{Temporal drift in ArcFace embeddings between consecutive frames for a real video (green) and its face-swap deepfake counterpart (red). Higher drift in initial frames and consistently higher overall drift in fake samples indicates temporal identity inconsistencies.}
    \label{fig:arcface_fig}  
\end{figure}

\section{Backgrounds}
Our method is based on observations of the temporal inconsistencies existing in various types of deepfakes, and these observations provide the foundation for the model architecture and detection strategies discussed in subsequent sections.

\noindent{\bf Phonemes Visemes Mismatch Pattern}. Lip-sync deepfake videos may have temporal inconsistencies due to the mismatch of phonemes and visemes \cite{Agarwal_2020_CVPR_Workshops}. To demonstrate this phenomenon, we selected a curated set of 14 phonemes based on their distinct articulatory features, high visual salience, and coverage of key phoneme classes important for audiovisual speech modeling and deepfake detection.
This subset spans a diverse range of phonetic categories:
\begin{itemize}
\item \textbf{Bilabials:} /p/, /b/, /m/ — require full lip closure and are visually distinguishable.
\item \textbf{Labiodentals:} /f/, /v/ — involve lip-to-teeth contact, producing clear articulatory movements.
\item \textbf{Alveolars:} /t/, /s/ — frequent in natural speech and involve rapid tongue and jaw movement.
\item \textbf{Velars:} /k/ — exhibit distinct mouth shapes with less lip movement but unique viseme cues.
\item \textbf{Approximants:} /w/, /\textipa{r}/ — require lip rounding or protrusion, easily detected in lip shapes.
\item \textbf{Vowels:} /i/, /\ae/, /o/ — include high front, low front, and mid back vowels, capturing open-mouth gestures.
\item \textbf{Postalveolar:} /\textesh/ — involves noticeable lip rounding and is visually distinct.
\end{itemize}

These phonemes are chosen to maximize the ability to detect mismatches between visual articulation and audio content, particularly in manipulated videos where viseme–phoneme alignment is disrupted. Phonemes with low visual distinguishability (e.g., glottals or unstressed vowels) and silences were excluded to avoid noisy or uninformative supervision.
We observe that lip-sync manipulations often fail to maintain alignment between speech articulation and lip movements. This discrepancy is mostly observed in labial,  labiodental, and vowel phonemes such as /p/, /b/, /m/, /f/, /v/, and /o/, which require precise lip closure or openness. 

As presented in Fig.~\ref{fig:main_fig}, in several lip-sync generated videos, the lip geometry failed to match expected phoneme articulations, particularly during high closure phonemes. These mismatch signals were consistently detected using MediaPipe-based lip geometry analysis.

\noindent{\bf Temporal Drift in Identity Embeddings}. We observe that for real videos, ArcFace-based identity embeddings transition smoothly across the frames. In contrast, fake videos, especially generated by face-swapping manipulation techniques, display sharp, irregular embedding shifts, suggesting that identity preservation across time is a useful indicator of face-swaps.

As seen in Fig.~\ref{fig:arcface_fig}, a plot of $\ell_2$ distances between consecutive ArcFace embeddings reveals that real videos maintain low and stable distances between 2-6 $\ell_2$ distance while fake videos exhibit sharp spikes at 8-12 $\ell_2$ distance in the beginning. This supports the use of identity drift as a reliable indicator for early-stage manipulation detection.

\section{Method}
\label{sec:method}

\begin{figure*}[t]
    \centering
    \includegraphics[width=1.0\textwidth]{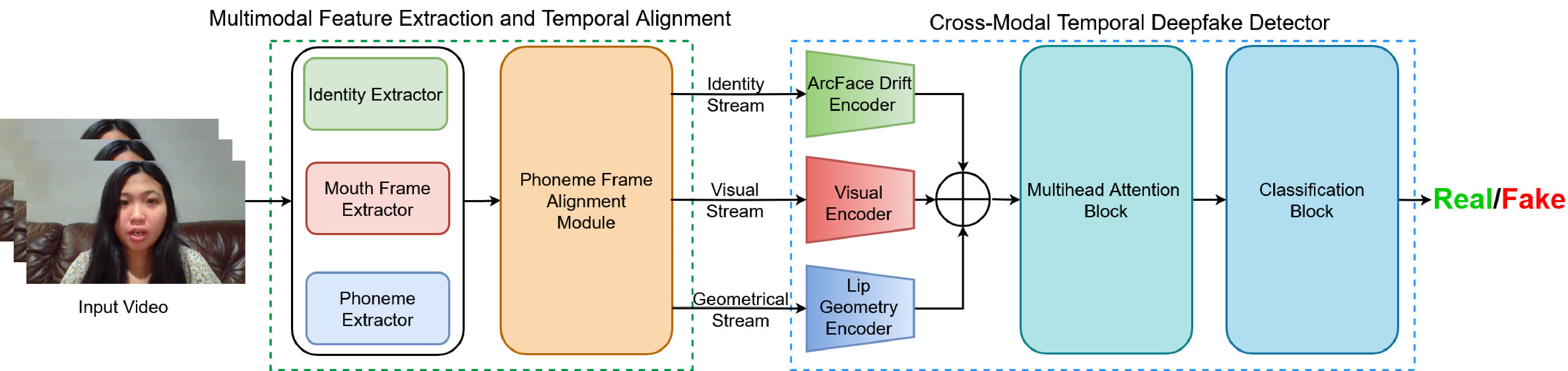} 
    \caption{End-to-end pipeline of our proposed PIA model. It consists of (1) Multimodal Feature Extraction and Temporal Alignment, and (2) Cross-Modal Temporal Deepfake Detector.}
    \label{fig:Model_arch}
\end{figure*}

Our proposed method integrates a multimodal feature extraction pipeline with a unified deepfake detection model that simultaneously analyzes phoneme articulation, lip geometry, viseme appearance, and identity cues. The model employs a 3D convolutional network \cite{tran2015learning} with a pre-trained EfficientNet-B0~\cite{tan2019efficientnet} backbone for extracting visual information from mouth-region images, along with a multihead attention \cite{vaswani2017attention} mechanism to efficiently  capture temporal and modality-specific relationships. Phonemes obtained by WhisperX~\cite{bain2022whisperx} and aligned with wav2vec2 \cite{xu2021simple}, are used as an active pre-processing filter to select temporally meaningful frames when phoneme articulation is visually significant. This architecture is designed to detect barely noticeable discrepancies caused by lip-sync and face-swap manipulations by studying cross-modal mismatches between audio and visual streams.
\subsection{Feature Extraction}
For each input video, we perform a structured pre-processing procedure to extract synchronized audio-visual and geometric features required for downstream deepfake detection. This process includes four key stages: audio extraction and phoneme alignment, visual feature extraction, facial identity embedding, and frame-level alignment.

\noindent{\bf Audio Extraction and Phonemes Alignment}.
We begin the process by extracting the raw waveform from every video, using FFmpeg \cite{ffmpeg} and resampling the audio to a 16 kHz mono-channel format in order to preserve uniformity among samples. Speech transcription uses the WhisperX \cite{bain2022whisperx} large-v2 model, producing word-level segments accompanied by accurate timestamps. The segments are subsequently transformed into sequences of International Phonetic Alphabet (IPA) phonemes via the phonemizer \cite{Bernard2021} package. We use a wav2vec2-based alignment model \cite{xu2021simple} to accurately match phonemes with the audio waveform, improving the phoneme timestamps at sub-word resolution. Ultimately, we associate each video frame with a corresponding phoneme label by interpolating its timestamp inside the matched phoneme intervals, therefore assuring synchronised frame-level phoneme annotation.

\noindent{\bf Visemes Feature Extraction}.
Each frame is analysed using MediaPipe \cite{lugaresi2019mediapipe} FaceMesh, which identifies 468 facial landmarks with exceptional spatial precision. To concentrate on articulatory dynamics pertinent to speech, we extract a subset of 27 lip-related landmarks that correspond to critical places along the outer and inner contours of the mouth.
We derive geometric descriptors from these locations to quantify lip movement and morphology. Lip height is the vertical distance between the central landmarks of both the upper and lower lips, whereas lip width is the horizontal distance between the left and right corners of the lips. Utilising these two measurements, we determine the mouth aspect ratio (MAR), defined as the ratio of lip height to lip width. 
\begin{equation}
\text{Aspect Ratio} = \frac{\text{lip\_height}}{\text{lip\_width} + \varepsilon}
\end{equation}
where $\varepsilon$ is a small constant (e.g., $1 \times 10^{-6}$) added to prevent division by zero.
This ratio functions as a crucial measure of mouth openness, with elevated values associated with vowel-like articulations and diminished values generally noted during lip closure or consonant production. These geometric indicators are especially valuable for discerning phoneme-specific articulation patterns and for recognizing discrepancies in deepfake videos. Bilabial phonemes such as /m/, /b/, and /p/ are anticipated to demonstrate low mouth aspect ratio (MAR) values owing to complete lip closure, but open vowels like /a/ or /i/ yield greater mouth aspect ratio (MAR) values due to vertical mouth extension.

\noindent{\bf Facial Identity Embedding}. We use the ArcFace \cite{deng2019arcface} model from InsightFace \cite{insightface2023} to obtain frame-level speaker identity representations. ArcFace is a state-of-the-art face recognition model that projects facial features into a 512-dimensional hypersphere space using an additive angular margin loss, which ensures high inter-class separability and intra-class compactness.
Each frame of the input video is independently processed by the ArcFace model to generate a 512-dimensional identity embedding. These embeddings encapsulate advanced facial characteristics, including skeletal structure, facial ratios, and expression-invariant traits, making them well suited to detect the slight yet discernible identity drift between frames caused by face-swap induced temporal anomalies. By examining the coherence of ArcFace embeddings across successive frames, the model can identify identity drift that may not be evident at the pixel level. These embeddings are retained for two key purposes:
(1) as one of the input modalities in the multimodal fusion model, where they are encoded and integrated with  viseme and geometric features, and
(2) for auxiliary supervision via a temporal consistency loss, which penalizes abrupt changes in identity features across adjacent frames and encourages the model to learn stable identity dynamics typical of real videos.

\noindent{\bf Multi modal Representation}. To train the model using frame-level multi-modal cues,  we construct a dataset in which each phoneme label is temporally aligned with its corresponding audio segment and drawn from a predefined set of 14 phonemes. For each phoneme in this set, five frames are uniformly sampled from the video instances where that phoneme appears.
For each of these frames, we extract three input streams: (1) viseme image crops, (2) identity embeddings generated using ArcFace, and (3) geometric descriptors computed from lip landmarks. Non-linguistic tokens, including silences, noise, and pauses, are omitted. Phoneme tokens are used to filter frames according to a carefully selected vocabulary of 14 visually and articulately distinct elements. This selection guarantees that only frames associated with prominent speech articulations are preserved for model training, hence improving the quality of visual, geometric, and identification aspects.

\subsection{Model Architecture}


Our multistream deepfake detection model integrates three modalities: (1) lip geometry descriptors, (2) viseme image crops, and (3) ArcFace identity embeddings. Each stream is encoded independently using a dedicated encoder, and the resulting features are fused via multi-head attention based pooling~\cite{vaswani2017attention} for final classification, as shown in Fig.~\ref{fig:Model_arch}. The ArcFace drift encoder is a multilayer perceptron that encodes the 512-dimensional ArcFace identity embeddings for each phoneme group. The visual encoder comprises a 3D convolutional network designed to capture spatio-temporal inconsistencies across the five-frame sequence associated with each phoneme. The resulting outputs are temporally averaged and subsequently processed by a pretrained EfficientNet-B0 backbone to extract higher-level visual representations. The lip geometry encoder is a multilayer perceptron that encodes the mouth aspect ratios computed from lip landmarks across frames.

To integrate multimodal features and summarize temporal dynamics, we concatenate modality-specific embeddings (geometry, visual, identity) into a unified feature:
\begin{equation}
\mathbf{f}_t \in \mathbb{R}^{3d}, \quad \mathbf{f}_t = \mathbf{g}_t \oplus \mathbf{v}_t \oplus \mathbf{a}_t
\end{equation}
The resulting sequence $\{\mathbf{f}_t\}_{t=1}^T$ is then summarized using multi-head attention pooling to obtain a global video-level representation.
Each learnable query attends over the temporal sequence to produce head-specific summaries, which are averaged to yield a compact representation:

\begin{equation}
\textstyle
\mathbf{z} = \frac{1}{H} \sum_{h=1}^H \sum_{t=1}^T \alpha_{h,t} \cdot \mathbf{f}_t'
\end{equation}
where $\alpha_{h,t}$ is the attention score assigned to each input frame \textit{t} by attention head \textit{h}.

\subsection{ArcFace Temporal Consistency Loss}

To enforce temporal coherence in the identity representation, we introduce an \textit{ArcFace Temporal Consistency Loss}, which penalizes abrupt or implausible changes in the ArcFace facial identities across consecutive frames.

Let $\mathbf{a}_1, \mathbf{a}_2, \ldots, \mathbf{a}_T \in \mathbb{R}^d$ denote the ArcFace embeddings for a video sequence of $T$ frames. For each adjacent pair of time steps $(t, t+1)$, we compute the cosine similarity:
\begin{equation}
s_t = \cos(\mathbf{a}_t, \mathbf{a}_{t+1}) = \frac{\mathbf{a}_t \cdot \mathbf{a}_{t+1}}{|\mathbf{a}_t| \cdot |\mathbf{a}_{t+1}|}
\end{equation}
We define the identity deviation as $1 - s_t$, which reflects the degree of identity shift between frames. To ignore irrelevant frames (e.g., during silence), we apply a binary mask $m_t \in {0, 1}$, where $m_t = 1$ indicates a non-silent frame.

The overall ArcFace temporal consistency loss is then defined as:
\begin{equation}
\mathcal{L}{\text{arcface}} = \frac{\sum_{t=1}^{T-1} (1 - \cos(\mathbf{a}_t, \mathbf{a}_{t+1})) \cdot m_t \cdot m_{t+1}}{\sum_{t=1}^{T-1} m_t \cdot m_{t+1} + \epsilon}
\end{equation}
where $\epsilon$ is a small constant added for numerical stability. This loss encourages smooth identity transitions in real videos and helps expose temporal inconsistencies introduced by manipulations in deepfakes.
The final loss is calculated as:
\begin{equation}
\mathcal{L}_{\text{final}} 
= \mathcal{L}_{\text{CE}} + \lambda \, \mathcal{L}_{\text{arcface}}
\end{equation}
where \(\mathcal{L}_{\mathrm{CE}}\) is the cross-entropy loss, and \(\lambda\) is a weighting coefficient on the ArcFace temporal consistency loss.

\section{Experiments}
\label{sec:experiment}

\begin{figure}[t]
    \centering
    \includegraphics[width=0.5\textwidth]{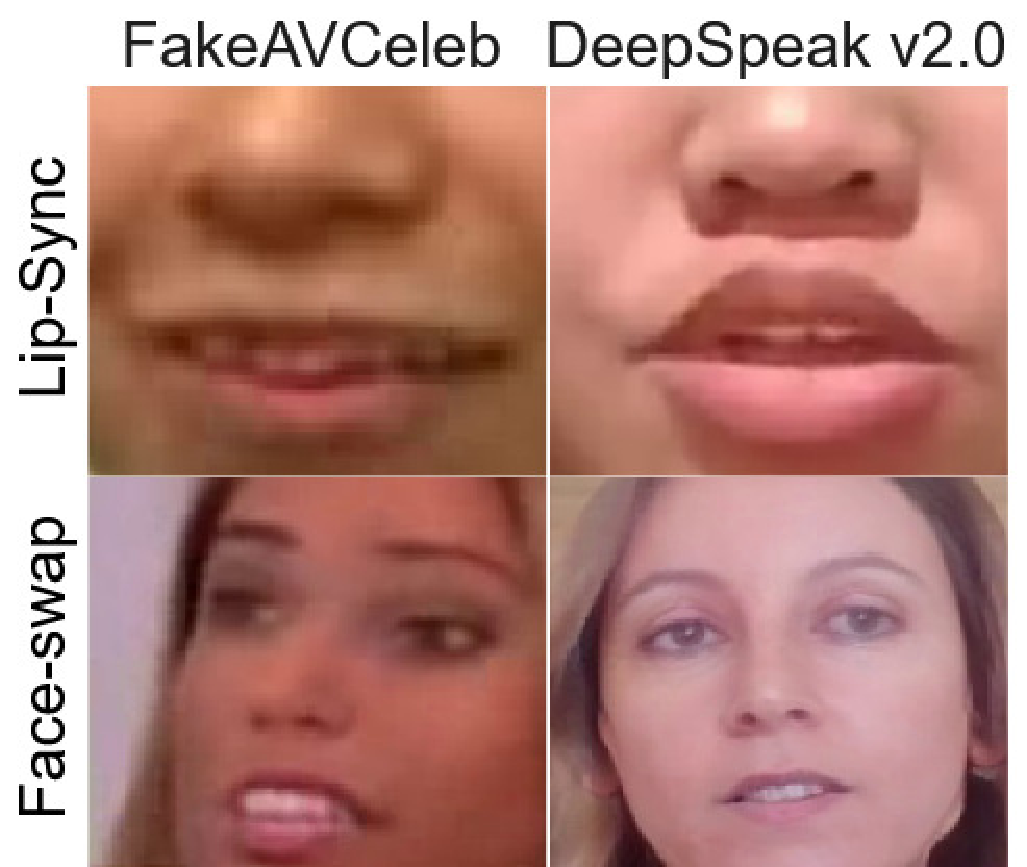} 
    \caption{Example frames from the FakeAVCeleb dataset (left) and the DeepSpeak v2.0 dataset (right). The top row shows frames from lip-sync deepfakes, while the bottom row presents frames from face-swap deepfakes. Notable differences in native resolution and visual quality are apparent. For comparison, face regions from FakeAVCeleb were enlarged to match the scale of DeepSpeak v2.0 samples.}
    \label{fig:deepspeakquality}
    
\end{figure}

\subsection{Experimental settings}
\textbf{Datasets:} In this paper, the experiments are performed on two datasets, namely  FakeAVCeleb  \cite{fakeavceleb} and DeepSpeak v2.0  \cite{deepspeak}. The FakeAVCeleb dataset\cite{fakeavceleb} consists of 20,000 samples, out of which 19,500 are deepfake videos and 500 are real videos of resolution $224 \times 224$. 
We partitioned the dataset into five categories similar to \cite{avff}:  
\begin{itemize}  
  \item FVRA-WL: FakeVideo-RealAudio-Wav2Lip \cite{wav2lip} 
  \item FVFA-FS: FakeVideo-FakeAudio-FaceSwap \cite{face-swap}
  \item FVFA-GAN: FakeVideo-FakeAudio-FaceSwapGAN\cite{fsgan}
  \item FVFA-WL: FakeVideo-FakeAudio-Wav2Lip \cite{wav2lip} 
  \item RVFA: RealVideo-FakeAudio
\end{itemize}

Following \cite{avff}, we used 70\% of the dataset to train and validate our model, and the remaining 30\% to test our model. Since our model is designed exclusively for fake-video detection, we exclude the RealVideo–FakeAudio category from both the training and test sets.

The DeepSpeak v2.0 dataset \cite{deepspeak} consists of 9,376 real videos and 7,209  deepfake videos, such as face-swapping, lip-syncing, and avatar-based fake videos. The real videos in this dataset are of two different resolutions, $640 \times 480$ and $1280 \times 720$. The fake videos are found with three different resolutions: $640 \times 480$, $512 \times 512$, and $1280 \times 720$. As provided by the dataset, the videos are split into training and testing subsets in an 80:20 ratio, respectively. We further divide the test set into three categories based on the type of deepfakes as Face-swap, Lip-sync, and Avatar. 

\noindent\textbf{Implementation Details:}
We use 14 distinct phonemes in the proposed architecture and construct the dataset by aligning phonetic alphabets extracted from WhisperX \cite{bain2022whisperx} with viseme crops and metadata on a per-frame basis. Mouth crops are resized to $112 \times 112$ pixels and normalised. The model is trained utilising cross-entropy loss with label smoothing and an auxiliary  ArcFace temporal consistency loss.
For all experiments we use the Adam optimizer with a learning rate of \(3\times10^{-4}\), weight decay of \(1\times10^{-5}\), and an  ArcFace temporal consistency loss coefficient \(\lambda=0.1\). For our multi-head attention module, we use 4 heads. We train our model for 25 epochs with a batch size of 16 using PyTorch 2.6.0 with CUDA 12.4.


\noindent\textbf{Evaluation Metrics:} The model is evaluated using three widely used metrics, including Average Precision (AP), Area Under the Receiver Operating Characteristic Curve (AUC) scores, and Accuracy (ACC) scores. We denote percentage points as \%-pts.

\subsection{Results}
\noindent\textbf{Performance on FakeAVCeleb Dataset.} In line with prior work \cite{avff}, we train our model on the FakeAVCeleb \cite{fakeavceleb} training split and evaluate it on the FakeAVCeleb test split.  As shown in Table~\ref{tab:FakeAV_performance}, our proposed method attains the highest results, achieving ACC and AUC scores of 98.7\% and 99.8\%, respectively, thereby surpassing all baseline models. For the PIA\_RVFA evaluation, the RVFA category is added back in the test set only and excluded from training. Although our model is designed solely for fake‐video detection, it achieves 98\% accuracy and a 98.2\% AUC inclusive of this category.

 In Table \ref{tab:Cross-Manipulation}, similar to \cite{avff}, we assess the model’s ability to generalize to videos manipulated by a method that was not included in the training set. We use four categories: FVRA-WL, FVFA-FS, FVFA-GAN, and FVFA-WL for this experiment. For each of the four categories, we partition the dataset into 70\% training and 30\% testing with no overlap. During training, we withhold one category entirely and use it solely for testing, thus rotating through all categories in turn.  It can be observed that our proposed approach is able to achieve the highest performance compared to the state-of-the-art model for all categories. For FVRA–WL, compared to LipForensics \cite{haliassos2021lips}, the best-performing baseline model in terms of AP, which achieved 97.8\% AP, our approach increases the performance by 2.1\%-pts, and when compared to AVFF \cite{avff}, the top model in terms of AUC with a 98.2\% AUC, we improve AUC by 0.9\%-pts. In terms of overall average performance(AVG-FV),  our method outperforms the best baseline model AVFF \cite{avff} by 1.4\%-pts  in AP and 0.3\%-pts in AUC.

\begin{table}[t]
  \centering
  \setlength{\tabcolsep}{4pt} 
  \begin{tabular}{@{}l c c c@{}}  
    \toprule
    Method & Modality & ACC(\%)  & AUC(\%) \\
    \midrule
    Xception \cite{Xception}            & V  & 67.9  & 70.5 \\
    LipForensics \cite{haliassos2021lips}& V  & 80.1  & 82.4 \\
    FTCN \cite{zheng2021exploringftcn}  & V  & 64.9  & 84.0 \\
    CViT \cite{cvit}                    & V  & 69.7  & 71.8 \\
    RealForensics \cite{RealForensics}  & V  & 89.9  & 94.6 \\
    \midrule
    Emotions Don’t Lie \cite{mittal2020emotions} & AV & 78.1 & 79.8 \\
    MDS \cite{chugh2020not}                     & AV & 82.8 & 86.5 \\
    AVFakeNet \cite{ilyas2023avfakenet}         & AV & 78.4 & 83.4 \\
    VFD \cite{cheng2023voice}                   & AV & 81.5 & 86.1 \\
    AVoID-DF \cite{yang2023avoid}               & AV & 83.7 & 89.2 \\
    AVFF \cite{avff}                            & AV & 98.6 & 99.1 \\
    \midrule
    PIA\_RVFA (Ours) & AV & 98.0  & 98.2 \\
    PIA (Ours)       & AV & \textbf{98.7} & \textbf{99.8} \\
    \bottomrule
  \end{tabular}
  \caption{\textbf{Performance on FakeAVCeleb Test Dataset.} We benchmark our method against baseline models on the FakeAVCeleb dataset using a 70\%/30\% train–test split. The best results are highlighted in bold.}
  \label{tab:FakeAV_performance}
  \vspace{-1mm}
\end{table}

\begin{table*}[ht]
  \centering
  \resizebox{\textwidth}{!}{%
  \begin{tabular}{l c *{5}{cc}}
    \toprule
    Method & Modality &
      \multicolumn{2}{c}{FVRA–WL} &
      \multicolumn{2}{c}{FVFA–FS} &
      \multicolumn{2}{c}{FVFA–GAN} &
      \multicolumn{2}{c}{FVFA–WL} &
      \multicolumn{2}{c}{AVG–FV} \\
    \cmidrule(lr){3-4} \cmidrule(lr){5-6} \cmidrule(lr){7-8} \cmidrule(lr){9-10} \cmidrule(lr){11-12}
    & & AP (\%) & AUC (\%) & AP (\%) & AUC (\%) & AP (\%) & AUC (\%) & AP (\%) & AUC (\%) & AP (\%) & AUC (\%) \\
    \midrule
    Xception \cite{Xception}           & V  & 88.2  & 88.3   & 92.3   & 93.5   & 67.6   & 68.5   & 91.0   & 91.0   & 84.8   & 85.3   \\
    LipForensics \cite{haliassos2021lips}  & V  & 97.8  & 97.7   & 99.9   & 99.9   & 61.5   & 68.1   & 98.6   & 98.7   & 89.4   & 91.1   \\
    FTCN \cite{zheng2021exploringftcn}      & V  & 96.2  & 97.4   & \textbf{100.0} & \textbf{100.0} & 77.4   & 78.3   & 95.6   & 96.5   & 92.3   & 93.1   \\
    RealForensics \cite{RealForensics}      & V  & 88.8  & 93.0   & 99.3   & 99.1   & 99.8   & 99.8   & 93.4   & 96.7   & 95.3   & 97.1   \\
    \midrule
    AV-DFD \cite{zhou2021joint}             & AV & 97.0  & 97.4   & 99.6   & 99.7   & 58.4   & 55.4   & \textbf{100.0} & \textbf{100.0} & 88.8   & 88.1   \\
    AVAD (LRS2) \cite{feng2023self}         & AV & 93.6  & 93.7   & 95.3   & 95.8   & 94.1   & 94.3   & 93.8   & 94.1   & 94.2   & 94.5   \\
    AVAD (LRS3) \cite{feng2023self}         & AV & 91.1  & 93.0   & 91.0   & 92.3   & 91.6   & 92.7   & 91.4   & 93.1   & 91.3   & 92.8   \\
    AVFF \cite{avff}                        & AV & 94.8  & 98.2   & \textbf{100.0} & \textbf{100.0} & 99.9   & \textbf{100.0} & 99.4   & 99.8   & 98.5   & 99.5   \\
    \midrule
    PIA (Ours)                          & AV & \textbf{99.9} & \textbf{99.1} & \textbf{100.0} & \textbf{100.0} & \textbf{100.0} & \textbf{100.0} & \textbf{100.0} & \textbf{100.0} & \textbf{99.9} & \textbf{99.8} \\
    \bottomrule
  \end{tabular}
  }
  \caption{\textbf{Cross-manipulation evaluation.} We evaluate the model’s performance by leaving out one category for testing while training on the rest, using a 70\%/30\% train–test split across four manipulation types (FVRA–WL, FVFA–FS, FVFA–GAN, FVFA–WL) on the FakeAVCeleb dataset. Each column heading indicates the held-out test category. The best results are highlighted in bold.}
  \label{tab:Cross-Manipulation}
\end{table*}

Our model's superior performance compared to state-of-the-art methods can be attributed to its capability to concurrently represent phoneme articulation based visual appearance, geometric consistency, and identification cues through a unified multimodal architecture. The use of cross-modal fusion, attention-based temporal pooling, and auxiliary ArcFace temporal consistency
loss enables the model to capture subtle spatiotemporal discrepancies that are often ignored by unimodal or weakly fused baselines.

\noindent\textbf{Performance on Deepspeak v2.0 Dataset.} Here we train and evaluate our model on the Deepspeak v2.0 dataset \cite{deepspeak}, which offers higher visual quality than FakeAVCeleb as shown in Fig.~\ref{fig:deepspeakquality}. We use the provided train/test split to evaluate our model. To the best of our knowledge, we are the first to benchmark on this Deepspeak v2.0; prior work \cite{yermakov2025unlocking} has only evaluated Deepspeak v1.0, which lacks avatar-based deepfakes. Their method \cite{yermakov2025unlocking} got an AUC score of 92.01\% on Deepspeak v1.0. Our proposed approach was able to achieve an AUC score of 98.06\% on  Deepspeak v2.0. This shows that our model is capable of detecting high-quality deepfakes as well. The results are shown in Table \ref{tab:pia_ablation}.

\setlength{\tabcolsep}{3pt}     
\begin{table}[h]
  \centering
  \begin{tabular}{lcccc}
    \toprule
    Dataset           & Lip-sync & Face-swap & Avatar & Global  \\
    \midrule
    PIA\_w\_ph\_w/o\_vi      & 68.44   & 62.12    &  64.73  & 65.57       \\
    PIA\_w\_ph\_w/o\_geom    & 98.31   & 91.56    & 96.77  & 96.49   \\
    PIA\_w\_ph\_w/o\_arc     &   98.64  &   95.99 & 96.68   & 97.02   \\
    PIA\_w\_ph           & 98.95   & 92.54    & 96.43  & 96.66   \\
    
    
    PIA\_w/o\_EB0     & 94.81   & 81.70    &  86.54 & 88.68      \\

    Vgg16\_w/o\_PIA      & 91.51   & 78.36    &  85.49 & 86.62      \\
     \midrule
    PIA               & \textbf{99.24}   & \textbf{96.47}   & \textbf{97.76}  & \textbf{98.06}   \\
    \bottomrule
  \end{tabular}
  \caption{Ablation analysis on Deepspeak v2.0 test set based on AUC (\%) scores. Global refers to the combined test set provided in the dataset. The best results are highlighted in bold.}
  \label{tab:pia_ablation}
  \vspace{-1mm}
\end{table}

\subsection{Ablation Analysis}
To evaluate the contribution of each component within our proposed framework, we present our ablation analysis in Table \ref{tab:pia_ablation}. We use the Deepspeak v2.0 dataset \cite{deepspeak} for ablation analysis,  since the Deepspeak v2.0 dataset has a higher visual quality as compared to the FakeAvCeleb dataset \cite{fakeavceleb} as shown in Fig.~\ref{fig:deepspeakquality}. 
All the models are trained on the training set of the Deepspeak v2.0 dataset. We report AUC scores for each ablation across the  Lip-sync, Face-swap, and Avatar test subsets, as well as for the combined test set referred to as Global in the Table \ref{tab:pia_ablation}.
We denote our full model as PIA. Here, “w/o\_vi” refers to excluding viseme image embeddings, “w/o\_geom” refers to excluding the lip geometry stream, “w/o\_EB0” refers to excluding the EfficientNet-B0 \cite{tan2019efficientnet} CNN backbone, and “w\_ph” refers to including the one-hot encoded phonemes data stream.






\noindent\textbf{Excluding Visemes (PIA\_w\_ph\_w/o\_vi)}
We train the model with the one-hot encoded phonemes and remove the visemes images input feature in order to assess the impact of visual cues on detection performance. It can be observed that the AUC falls by 30.8\%-pts, 34.35\%-pts, 33.03\%-pts, and 32.49\%-pts for  Lip-sync, Face-swap, Avatar, and Global test subsets, respectively.
These results highlight the critical role of visual image cues from the lip region in capturing the inconsistencies.


\noindent\textbf{Excluding Lips Geometry (PIA\_w\_ph\_w/o\_geom)}
In this experiment, we train the model with the one-hot encoded phonemes and remove the lip geometry stream to evaluate the contribution of geometric features in deepfake detection. By removing this component, we assess the model’s reliance on temporal lip shape variations for capturing subtle inconsistencies in lip-sync manipulations. The results show a slight drop in AUC of 0.93\%-pts, 4.91\%-pts, 0.99\%-pts, and 1.57\%-pts for  Lip-sync, Face-swap, Avatar, and Global test subsets, respectively. This suggests that lips geometry provides complementary information.

\noindent\textbf{Excluding ArcFace Embeddings (PIA\_w\_ph\_w/o\_arc)}
In this experiment, we train the model with the one-hot encoded phonemes and exclude the ArcFace identity embeddings. By eliminating this stream, we assess the model's capacity to identify  temporal inconsistencies without identity-based signals. Here, the AUC dropped by 0.6\%-pts,  0.48\%-pts,  1.08\%-pts, and 1.04\%-pts  for  Lip-sync, Face-swap, Avatar, and Global test subsets, respectively.

\noindent\textbf{Including One-Hot Encoded Phonemes (PIA\_w\_ph)}
In this ablation, we train the model by introducing the one-hot encoded phonemes along with cross-modal attention applied to ArcFace facial embeddings, without removing any other module. These one-hot encoded phoneme features are fused with visual appearance and lip geometry streams to form the phoneme-infused baseline. Here the AUC dropped by 0.29\%-pts, 3.93\%-pts, 1.33\%-pts, and 1.40\%-pts for Lip-sync, Face-swap, Avatar, and Global test subsets, respectively.
These results suggest that while phonemes are useful for selecting relevant frames with aligned viseme images, lip geometry, and identity features, their inclusion as a fused input stream may introduce noise, thereby limiting the model’s discriminative capacity.


\noindent\textbf{Excluding EfficientNet-B0 (PIA\_w/o\_EB0)}
Here we train the model with frozen pretrained RESNET-18 \cite{he2016deep} embeddings in place of the EfficientNet-B0 \cite{tan2019efficientnet} backbone module to assess the impact of fixed visual representations on model performance.
By replacing EfficientNet-B0  CNN backbone in training with frozen pretrained RESNET-18 embeddings, we see the AUC dropped by 4.43\%-pts, 14.77\%-pts, 11.22\%-pts, and 9.38\%-pts for Lip-sync, Face-swap, Avatar, and Global test subsets, respectively.

\noindent\textbf{Using Vgg16 CNN model (Vgg16\_w/o\_PIA)}
In this experiment, we replace our PIA detection module with a simple VGG16 \cite{simonyan2015very} CNN architecture to assess the significance of phonemes, visemes, geometric, and identity cues for deepfake detection. Replacing our architecture with a simpler model resulted in the AUC dropping by 7.73\%-pts, 18.11\%-pts, 12.27\%-pts, and 11.44\%-pts for Lip-sync, Face-swap, Avatar, and the Global test subset, respectively.

From this analysis, we observe that integrating visual, geometric, and identity cues extracted using phonemes with an EfficientNet-B0 CNN backbone provides the most robust model to detect deepfakes.

\section{Conclusion}
In this paper, we offer PIA (Phoneme-Temporal and Identity-Dynamic Analysis), an innovative, unified multimodal technique for audio-visual deepfake detection. PIA concurrently models phoneme articulation, visual features, geometric consistency of lips, and identity indicators to identify subtle temporal and cross-modal inconsistencies. Our approach attains state-of-the-art performance, exhibiting robust generalisation in cross-manipulation settings. PIA demonstrates exceptional performance on the high-resolution DeepSpeak v2.0 dataset, demonstrating its resilience in authentic and high-fidelity deepfake contexts. 

Although our method achieves strong results on videos at the specific resolutions used for training, its generalization remains limited to those conditions, additional fine-tuning is needed to perform reliably on videos with different resolutions. In addition, due to our reliance on WhisperX and wav2vec2 for phonetic alignment, the model has been restricted to English-language inputs. Lastly, our approach is designed for scenarios involving fake videos and may not be applicable to instances of RealVideo-FakeAudio  (RVFA), where the visual elements are authentic but the audio has been manipulated.

The possible areas of extension to our work include generalization capabilities of our model to video resolutions beyond the training data,  by augmenting the training data with varied resolutions. Another significant direction is addressing RealVideo-FakeAudio (RVFA) cases. We intend to extend the model's capability to detect audio-only forgeries as well by introducing modality-specific anomaly detectors. Furthermore, to enhance multilingual deepfake detection, we intend to substitute the English-centric phonetic alignment with multilingual voice representations, such as those obtained by multilingual automatic speech recognition(ASR) speech-to-text models.

\noindent\textbf{Acknowledgment.} This work is supported by the Center for Identification Technology Research (CITeR) and the National Science Foundation under Grant No. 1822190.

{
    \small
    \bibliographystyle{ieeenat_fullname}
    \bibliography{bibmain}
}

\end{document}